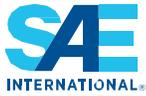

# Developing a Safety Management System for the Autonomous Vehicle Industry

**David Wichner**  Waymo, SAE ORAD Committee

**Jeffrey Wishart**  Science Foundation AZ/AZ Commerce Authority, SAE ORAD Committee

**Jason Sergent**  dss+

**Sunder Swaminathan**  Arizona State University

## Abstract

Safety Management Systems (SMSs) have been used in many safety-critical industries and are now being developed and deployed in the automated driving system (ADS)-equipped vehicle (AV) sector. Industries with decades of SMS deployment have established frameworks tailored to their specific context. Several frameworks for an AV industry SMS have been proposed or are currently under development. These frameworks borrow heavily from the aviation industry although the AV and aviation industries differ in many significant ways. In this context, there is a need to review the approach to develop an SMS that is tailored to the AV industry, building on generalized lessons learned from other safety-sensitive industries. A harmonized AV-industry SMS framework would establish a single set of SMS practices to address management of broad safety risks in an integrated manner and advance the establishment of a more mature regulatory framework. This paper outlines a proposed SMS framework for the AV industry based on robust taxonomy development and validation criteria and provides rationale for such an approach.

## Introduction

Regulators and safety investigation bodies in the transportation industry have either implemented or strongly advocated for companies to adopt robust SMS. The automated driving system (ADS)[1]-equipped vehicle (AV) industry has responded positively to this advocacy and developed proposals for an AV industry SMS framework [1] [2] [3]. There are also many safety-sensitive industry SMS frameworks that the AV industry can learn from, with many lessons to learn. Design of an effective AV SMS must address its novel combination of industry sectors, full value chain of functions, and the reality of a dynamic and maturing regulatory environment. This paper distills many of those lessons and suggests how an AV industry SMS framework might incorporate those insights.

This paper outlines an enhanced SMS standard framework for the AV industry based on lessons learned from other safety-sensitive industries, as well as a rationale that applies taxonomy development and validation criteria to improve the overall integrity of the framework's design. This model is offered as a proposed framework for organizations performing various functions in the AV-industry value chain, and may also provide insights to other industries that may be vulnerable to management factors not included in their SMS frameworks that can contribute to organizational accidents.

## *Understanding SMS*

An SMS[2] is a formal, leader-sponsored, organization-wide framework to help manage safety risk associated with all life cycle stages of a company's products or services. An industry SMS Recommended Practice or Standard is a consensus document that details a comprehensive, systematic, but generalized framework of management processes that helps companies identify safety-related hazards, reduce risk through effective management of safety controls, and establish capabilities that drive continual improvement of broad, interfacing, safety-related activities. Emphasis is placed on improving how decisions are made, establishing proactive processes, involving the workforce, and implementing internal controls to assess compliance and drive continual safety improvement.

Discussions about SMS began as early as the 1970's [4, p. 19]. As regulatory requirements became increasingly complex with advancing industrial technologies, the demand on regulators became too burdensome to be efficient or effective [4, p. 123]. SMS was seen as an important means to balance safety accountability between regulators and industry. Public pressure for better safety management was also driven after high visibility catastrophes, such as the Bhopal, India chemical industry tragedy in 1984. The establishment of the Center for Chemical Process Safety (CCPS) and its SMS framework called Risk Based Process Safety framework is one example of how an industry responded to pressure to improve its safety performance with an industry standard SMS approach [5].

---

1 In SAE J3016 [35], ADS refers to driving automation systems that are Levels 3-5.

2 For the purposes of this paper, the term SMS serves as a singular reference to any industry's set of safety management frameworks, standards, or recommended practice. This is regardless of the formal title of the framework. For example, while The American Institute of Chemical Engineers' (AIChE) Center for Chemical Process Safety (CCPS) Risk-Based Process Safety guidelines are not formally called an SMS, the authors of this paper will refer to it and its peer industry's safety management frameworks as SMSs.

Within transportation, SMS was first recognized in the Maritime industry when the International Maritime Organization's (IMO) adopted the International Management Code for the Safe Operation of Ships (ISM Code) in 1993, made mandatory in 1998 [6]. SMSs emerged in aviation when the International Civil Aviation Organization (ICAO) published its first Safety Management Manual (SMM) in 2006 [7]. The Federal Aviation Administration (FAA) published its first Advisory Circular outlining guidance for SMS for aviation service providers, such as air carriers and air taxi services [8] in 2006 as well. In 2010, the U.S. Congress mandated the FAA develop a rule requiring all U.S.-based airlines, regional air carriers, and cargo operators to implement an SMS [9]. In 2015, a final SMS rule was issued for larger air carriers, then expanded to cover certain aircraft manufacturers and smaller air operators in 2024 [10] [11]. In 2011, the U.S. Department of Transportation (DOT) asked each Operating Administration, such as the Federal Railroad Administration (FRA) and Federal Transit Administration (FTA) to incorporate principles of SMS into safety oversight activities of the transportation community [1]. As of 2024, each Operating Administration has adopted an SMS rule [12] with some variation on these four broadly common SMS components: Safety Policy, Safety Risk Management, Safety Promotion, and Safety Assurance [13].

Yet it is critical to remember that while the aviation and related transportation SMS regulations are widely recognized, other industries had previously developed their own SMS industry consensus standards. In fact, the concept of "SMS did not emerge in aviation and appeared long before it was adopted by aviation..." [4, p. 85]. A list highlighting samples of industries which have adopted their own form of safety management standards can be found in Appendix A and include nuclear, occupational safety, and chemical manufacturing.

### *Overview of SMS effectiveness*

Accident statistics are often seen as an indicator of successful management of safety. In that light, SMSs appear to be strongly correlated with accident reductions and have been a positive influence in aviation safety. According to the International Air Transport Association (IATA), between 2015 and 2023 the industry reported a 44% reduction in total events meeting the formal accident criteria compared to years prior to SMS implementation (i.e., 2005-2014) [14]. Similarly, between 2015 and 2021, the NTSB reported a 10% decrease in accidents per 100,000 flight hours [15]. Even with these rates improving, NTSB accident investigations continue to reveal that, in many cases, if effective SMS programs had been in place, they could have prevented additional loss of life and injuries [3].

According to the American Chemistry Council (ACC) [16], their member companies have seen broad improvements in safety performance since adopting relevant safety management practices. The ACC reports: 1) a reduced recordable injury and illness incidence rates by 20 percent from 2010 to 2022; 2) a reduced number of chemical distribution incidents among member companies by approximately 24% from 2017 to 2021; and 3) a reduced number of process safety incidents (unplanned events arising from the manufacturing process resulting in a product release, fire, explosion, injury, or community event) from 282 to 189 between 2011 and 2022. The industry itself attributes at least some of these improvements to a concerted effort to adopt broad industry safety management practices published by the American Petroleum Institute [17].

The Japanese High-Speed Railway (HSR) is another case which appears to demonstrate how SMS has been a positive contributing factor to its outstanding safety record of zero passenger fatalities [18]. In addition to reliance on advanced methods for management of complex socio-technical systems such as STAMP/STPA [19], SMS and its Safety Risk Management processes are also credited with having a foundational and positive effect.

These three industries give credit for improved safety records attributed, at least in part to the development and adoption of safety management frameworks. However, attention should also be given to areas where SMS practices may not have had a significantly positive impact on safety. Recent high-profile incidents involving Boeing's safety performance may be a good example in the aviation industry. While the specifics of these incidents is not the focus of this paper, they do present an interesting case study in the performance of SMS.

As previously mentioned, it was not until 2024 that a formal aviation SMS rule was applied to aircraft manufacturers, even though Boeing had previously committed to implement an SMS in 2019 following the 737-MAX accidents [20]. The point here is that Boeing's safety and quality issues do not necessarily demonstrate a failure of SMS principles per se. Rather, these circumstances raise questions about why formal SMS regulations were not applied earlier to manufacturers such as Boeing, and whether Boeing's recent safety performance might have been better had the FAA required SMS adoption across the entire aviation value chain from the outset, instead of initially limiting the regulation to service providers.

In terms of general industry occupational safety, the U.S. Occupational Safety and Health Administration (OSHA) reported a fatal workplace injury rate of 3.7 per 100,000 full-time equivalent workers for 2022, up from 3.5 in 2011 [21]. This trend belies efforts to adopt Occupational Health and Safety Management System (OHSMS) practices. This trend differs significantly from the apparent successes in the other industries cited, but unfortunately research shows "that there is room for improvement in understanding the effectiveness of OSH interventions" with respect to workplace fatality rates [22] [23].

Unlike the aviation industry, statistics from the National Transit Database (NTD) indicate that accident rates in the public transit sector have also increased between 2012 and 2022 (from 389 reportable safety events per 100 million vehicle revenue miles (VRM) in 2012 to 512 reportable safety events per 100 VRMs in 2021) [24].

In practice, implementation of SMS in the public transit industry has required the adoption of multiple SMS requirements and structures. Multi-modal transit agencies, for example, operating subway, commuter rail, and bus transit services, must meet the SMS requirements of both the Federal Transit Administration (FTA) and those of the Federal Railroad Administration (FRA) [25] [26]. The authors surmise that inconsistencies in terminology and requirements may aggravate transit agency's ability to implement an effective SMS.

Of special note regarding the effectiveness of the aviation SMS overall, a review of a recent exhaustive and extremely well researched analysis of its context, rationale, origins, functions, strengths, and vulnerabilities highlights opportunities inherent in the aviation SMS framework [4]. In summary, this analysis of the design and implementation of the aviation SMS framework highlights design deficits in the framework itself that serve as a warning to a new industry seeking to adopt an SMS that it should be more inclusive of benchmarks it uses to design its SMS framework.

In response to criticisms that implementing an SMS is an administrative burden without a salient improvement to safety, former Chair of the NTSB Robert Sumwalt stated "Just forget SMS. Instead, think of it this way: The things that are part of a fully functioning SMS are the very things a professionally run aviation provider should be doing in the first place." [27]. This comment can equally apply to any other safety-sensitive business or industry. At this stage in the history of safety, a robust industry safety management framework should be considered consistent with the standard of care for a company's management of its safety practices. In this light, challenges to improving the design and adoption of a robust SMS should simply be seen as hurdles we must overcome.

## *Comparing industry SMS frameworks*

If we have a general understanding of where SMS comes from, as well as how effective we might expect them to be, what elements should the AV industry draw from? The body of research into management factors which may contribute to accidents is extensive. Bieder [4] researched and referenced hundreds of frameworks, journal articles, conference presentations, and foundational texts in her socio-historical analysis of the aviation SMS. One such reference states "...coupling a more powerful socio-technical understanding of complex systems, their operation, and their risks to safety requires also that we increase the scale, scope and time frame of safety management itself." [28]. This analysis suggests a more robust SMS framework should go beyond the basic four element model and include management functions that have been found to contribute to accidents. Although not comprehensive, Appendix A includes a list of SMS frameworks which represent this broader perspective.

By comparing and contrasting these different frameworks in detail, the AV industry can identify important safety management functions established in other industries and organize them in a way that is more logical for the AV industry and better fits within its own context.

This set of frameworks and their constituent elements were compared to identify commonalities and differences. The comparison reveals similarities in substance, but many significant differences in scope, hierarchy, and terminology. Figure 1 is a simplified side-by-side comparison of the tables of contents of the aviation SMS, ISO Integrated Management System Standard, API RP 1173, and CCPS' Risk Based Process Safety frameworks [11] [29] [17] [30].

| Aviation | ISO (harmonized) | Pipeline | Chemical |
|---|---|---|---|
| Safety Policy | Context of the organization | Leadership & Management Commitment | Commitment to Process Safety |
| Safety Risk Management | Leadership | Stakeholder Engagement | Understanding Hazards & Risks |
| Safety Assurance | Planning | Risk Management | Manage Risk |
| Safety Promotion | Support | Operational Controls | Learn from Experience |
|  | Operation | Incident Investigation, Evaluation & Lessons Learned |  |
|  | Performance evaluation | Safety Assurance |  |
|  | Improvement | Management Review and Continuous Improvement |  |
|  |  | Emergency Preparedness & Response |  |
|  |  | Competence, Awareness & Training |  |
|  |  | Documentation & Record Keeping |  |

Figure 1: High level comparison of SMS elements

General observations from this comparison include:

- Some elements listed in one framework may be embedded deeper within the hierarchy of other frameworks.
- Frameworks address safety management practices in different ways, such as being distributed or centralized in the structure.
- Elements appear fully expressed in some frameworks, while in others they are treated in a cursory manner, or not at all.
- Some SMS include elements or functions that are also regulated, while others do not.
- Frameworks often define similar elements with different terms and scope.

One major conclusion that can be drawn is that the ecosystem of global industry SMS frameworks and elements is diverse with variability in how they may or may not have applied taxonomy development and validation criteria discussed below. Applying insights from these these observations can help to improve the proposed AV SMS framework address these gaps.

One example of a common gap in most SMS frameworks is the lack of a systemic structure for the management of safety culture. Although the aviation industry strongly advocates for creation of a positive safety culture [33] [7], there is no mention of safety culture in the FAA's SMS regulation [11], let alone how a company might manage it effectively. Additionally, in the current version of the AVSC SMS Information Report [34], safety culture is seen in short descriptive phrases and seen as an outcome, rather than a process to be managed in its own right. In contrast, the nuclear power and chemical industries have invested significant effort in defining safety management frameworks and specific attributes of a healthy safety culture [31] [32].

Similarly, implementation effectiveness of the AV SMS can be improved by looking at how elements are used in context with each other. For example, in the AVSC Information Report, a Safety Reporting System is seen as a function of, and subordinate to Safety Policy. The stated rationale for this is that a "Safety Reporting System is used to bring safety hazards to the attention of organizational leadership, thus supporting the Safety Policy." [34]. This claim can easily be made for any of the provisions of an SMS. However, in the definition itself, a Safety Reporting System is axiomatically a means to identify safety concerns, which is an essential core function of Safety Risk Management (SRM). This is an example of logical and functional disassociation of related functional elements that may create some confusion of the purpose of a function.

It is highly doubtful that any company would fail to adopt a key element like a Safety Reporting System because of this error. However, there are quite a few such issues like this which can make the overall system more difficult to comprehend and manage efficiently, thereby impacting ease of adoption in an industry unfamiliar with the concept of an SMS.

To help develop a harmonized AV industry SMS and avoid these design challenges we must first understand the context of the AV industry. Several methods can then be used to 1) include diverse safety-related management functions, and 2) ensure they are organized in a comprehensive, logical, and easy to use SMS framework.

## *Overview of the AV industry*

While the AV industry has already seen commercial deployment of the technology, the industry is still exploring new applications and business models. The current leading business models for AV deployment at SAE Levels 4 and 5 in particular [35] include a novel combination of industrial sectors including: automotive vehicle and equipment manufacturing, Artificial Intelligence (AI)-intensive software, and transportation services and operations. As a result of this confluence of industry sectors, there are unique challenges with integrating how to manage the safety risks of development, testing, and commercial deployment of AVs. The design of an SMS for this industry must effectively address this novel combination of industry sectors, full value chain of functions, and the reality of a dynamic and maturing regulatory environment.

The AV industry has made significant efforts to update existing automotive safety standards (e.g., ISO 26262 and ISO 21448) [36] [37], as well as develop new safety standards, best practices, and information reports (e.g., ANSI/UL 4600, SAE J3018, SAE J3237, and AVSC-I-01-2024) [38] [35] [39] [34]. These documents provide useful guidance on how ADS should be developed, tested, and deployed to a reasonable level of safety. However, this complex combination of industry sectors also creates novel challenges that have not yet been faced by these sectors individually, including fundamental differences, gaps, and potential conflicts in regulatory environments, as well as safety best practices and terminology.

Participants in the AV industry coming from each of these different industry sectors may not have a common understanding of what safety means in terms of regulated functions, industry standards, key terminology and definitions, or even the type of risk factors that may be present. For example, automotive safety experts may not have in-depth knowledge of machine learning or transportation services and operations safety functions. Transportation services and operations experts may not have in-depth knowledge of automotive engineering and manufacturing safety practices. These differences in specialized industry domain knowledge pose a unique challenge to the AV industry overall. Therefore, the development of an SMS for the AV industry depends on establishing and using common safety terminology, definitions, and practices. This can only occur if industry participants from these various industries acknowledge the limitations of their historic practices within the new, cross-sector industry of AV development, test, and deployment.

This notion is at the very heart of the approach proposed here to assess and harmonize a broad sample of safety management frameworks to develop a new AV industry SMS. As a new industry, careful consideration must be given to mitigate the vulnerabilities of organizational risk factors which may have been identified by other, apparently unrelated industries. After a survey of several SMS frameworks, it is the contention of the approach offered here that companies in different industries which create and offer different products and services are still vulnerable to common management factors that can contribute to organizational accidents.

When considering this inclusive approach to design an AV SMS, it is important to guard against perspectives that other industries' safety management practices as not applicable to the AV industry because they were "not invented here". Valid lessons can be learned from safety practices that originate far afield. At the same time, an effort must be made to ensure that a proposed AV industry SMS addresses questions of design relevancy, inconsistencies, and are not simply copied without appropriate evaluation and adaptation.

**The AV industry value chain**

Concepts such as how safety is managed between contracted partners is often part of SMS frameworks. This implies an obligation of companies to assure effective management of safety across up and down their value chain. So, while an SMS is implemented by a single company, the value chain of an industry is arguably the highest order consideration for informing the design of an SMS at the industry level.

We can see this in the widely-accepted Quality Management System (QMS) standard for the automotive manufacturing sector which outlines quality practices for the entire value chain [40]. It is therefore reasonable for the AV SMS to include guidance for safety management practices that apply to all relevant value chain functions in addition to the technical safety standards referenced earlier.

Figure 2 illustrates a notional view of the AV industry value chain. It shows that the engineering and manufacturing phases of the value chain are typically in the domain of the automotive OEMs, operations are the domain of owner-operators (fleet and individuals), while disposition is typically a shared responsibility of the two entities. For their part, AV companies may perform any number of functions across the full AV industry value chain as shown here. This is also a way of visualizing the inherent challenges discussed earlier that are present when different industries are brought together to create a new product. How can an industry effectively manage safety if parts of the value chain are not integrated into the same SMS framework. This observation is reflected in our previous discussion about Boeing and why SMS regulations were not applied to manufacturers earlier.

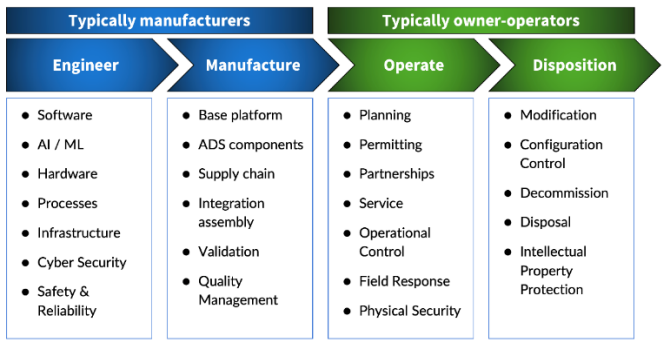

*Figure 2: AV value chain and sample functions*

The AV industry is not simply engineering and manufacturing the vehicle and its technologies to deliver safe Dynamic Driving Tasks (DDT). The value chain also includes commercial deployment functions such as ride-hailing, trucking, fleet management, remote and in-field assistance, and other transportation services and operating functions that have not been the historic domain of either automobile OEMs or the AI-intensive software sector. An AV SMS should provide generalized guidance on safe practices for all functions of the entire AV value chain from the outset.

**Supporting an evolving regulatory environment**

The AV industry and its value chain differs in other important ways from other transportation sectors. As an example, aviation safety in the U.S. is regulated by a single federal agency across all value chain functions, the FAA. In contrast, safety regulations for the AV industry are separated by function across federal, state, and even local jurisdictions. They are also still maturing even before considering the rapid pace of change. Recognizing this current state, a robust AV SMS would help industry stakeholders have a degree of confidence that consistent and comprehensive safety management practices are available for companies to implement. However, the effect of this landscape remains a significant open question and one where more research is needed.

*SMS and the AV industry*

SMS is an important tool to proactively and efficiently manage safety by enabling improved oversight of a company's safety-related processes. For the AV industry, this is important in light of the limited and maturing regulatory structure. As the industry evolves, an SMS and standardized terms, definitions, and taxonomies can improve transparency. This benefit serves to support consistent interpretation of contract obligation and audits of companies' safety processes, protect sensitive intellectual property, all the while giving stakeholder like the public, regulators, and contract partners a 'Rosetta stone' of normative practices to compare.

Beyond historic reasons to adopt SMS, such as catastrophic accidents and constraints on regulatory oversight, the AV industry may also have new drivers for adoption. First, this new industry has a limited deployment safety record, a low level of consumer control of outcomes, and a dynamic and maturing regulatory structure. As a result, the authors surmise that SMS in the AV industry may be driven by public and regulatory concerns over the possibility of a series of high-profile smaller accidents across the industry or within a single company. Regulators and investigating bodies have also stated that adoption of SMS is an important safety tool that should be employed in every transportation domain [41] [3].

Consistent with these drivers, the European Parliament has established requirements for AV manufacturers to establish and obtain third-party certificate of compliance for their SMS in its uniform procedures and technical specifications for the type-approval of the ADS of AVs [2]. Additionally, as demonstrated by documents such as AVSC's Information Report on SMS [34], the AV industry itself is committed to develop its own AV SMS framework.

However, each of these efforts borrows heavily from and extends little beyond the aviation/transportation SMS framework of the four elements described earlier. Even though they are not formally called SMS, the broad set of safety management frameworks found in industries beyond aviation are nevertheless an SMS and include best practice guidance on organizational factors that have contributed to accidents. As a result of their robustness and potential applicability to any company with safety-sensitive products and services, the authors believe these frameworks should be considered when designing a new SMS for the AV industry.

Because of the novelty of such an integration effort, special attention must be given to the design of other safety frameworks to better understand the lessons they offer. This is particularly true given the significant variations that exist between these standards and how they are applied based on specific industry needs. Such variations can be found in terminology, framework structure (number, type, and hierarchy of elements), as well as areas of focus or safety domain, i.e., product, public, workforce, and environmental safety. Ultimately, we should be able to answer which elements should be included in a new AV SMS and how they are arranged.

**Current SMS efforts in the AV industry**

The efforts outlined below outline the AV industry's efforts to develop its own SMS standard.

*AVSC SMS Information Report*

SAE ITC's Automated Vehicle Safety Consortium (AVSC) developed an initial SMS framework which the authors believe can serve as a valid starting point for any AV company [34]. It was intended to provide basic guidance to organizations on elements AV companies might use to assess and manage safety risks, evaluate effectiveness of risk control strategies, and establish and promote safety policies and objectives for organizations testing and evaluating SAE Level 4 and 5 vehicles. The AVSC SMS Information Report framework is consistent with the common SMS components of DOT-promulgated SMS rules: Safety Policy and Objectives (SPO), Safety Risk Management (SRM), Safety Assurance (SA), and Safety Promotion (SP).

*UNECE and the European Commission*

The United Nations Economic Commission for Europe (UNECE) established the World Forum for Harmonization of Vehicle Regulations to define requirements under the Intelligent Transport Systems (ITS) informal working group to focus on automated driving (ITS/AD). This group has formalized requirements for an AV industry SMS in their Informal Document GRVA-18-50 [42]. In close alignment, the European Commission published regulations [2] which also largely reflects the four elements of the aviation SMS framework.

*SAE International On-Road Automated Driving (ORAD) Committee*

SAE International ORAD Committee established a Task Force to develop a broad-based SMS. The first document the Task Force is developing is a Recommended Practice which expands upon the AVSC SMS Information Report. It is common for AVSC documents to prompt more formal technical standards efforts under SAE committees such as ORAD. The ORAD effort proposes to develop a more comprehensive SMS framework, harmonizing elements from a broad range of industries. This approach is driven by the novel fusion of industry sectors, the maturing regulatory environment, and the expansive and dynamic nature of AV value chain and business models.

As such, ORAD is pursuing an SMS framework that goes beyond the four elements used in other AV SMS frameworks. The Recommended Practice under development includes safety culture management as the first element; safety planning; safety governance; safety risk management; and finally, safety life cycle management to ensure inclusion of elements that cover all the phases in the AV value chain.

*ANSI / UL 4600*

ANSI and UL partnered to develop the Standard for Safety for the Evaluation of Autonomous Products, a safety case[3]-based approach to assess the safety of AVs and demonstrate through documentation and performance that AVs (SAE Level 4 and 5) are acceptably safe [38]. This document is structured with prompt elements that outline topics to be addressed while developing the safety case.

ANSI/UL 4600 offers some elements that appear to overlap with a comprehensive SMS framework, such as an approach to identify the role of safety culture in the organization and assess the risk, implement mitigation strategies, and evaluate the effectiveness of the mitigation strategies. Additionally, ANSI/UL 4600 provides recommendations covering the safety aspect across the entire life cycle of AVs. However, the document does not lay out a comprehensive SMS. Rather, it establishes an approach to develop and assess a case for the reasonable safety of the product. Nevertheless, the elements of ANSI/UL 4600 are an important complement to a fully harmonized SMS framework A safety case is an important product of a robust SMS. While safety assurance is much broader than the safety case, the safety case utilizes safety assurance techniques to justify that an AV system is safe enough to be deployed on public roads without a human driver, and includes evidence to support that determination. More research should be conducted on the relationship between SMS and the Safety Case.

*Summary: The case for a new approach*

By evaluating results of the comparative analysis across different industries and the assessment of the current state of AV industry SMS efforts, we see there is opportunity to develop a broader framework. Yet, given the fact that the AVSC SMS [34] and EU regulations [2] are already published, one can ask why the aviation-based SMS is insufficient. A recent and extensive evaluation drew an important conclusion as to the completeness and efficacy of the aviation SMS framework. "As an approach to safety, the [aviation] SMS turns out to have a number of limitations. Although it may help in identifying major and relatively stable issues, all the dynamic aspects and contingencies go through the SMS net without being spotted or addressed." [4, p. 79]. The proposed approach is intended to establish an SMS framework that can be applied to the AV industry in a way that closes as many of these gaps and inconsistencies as practicable.

## Approach

*System safety principles and SMS design*

A principle of system safety is to evaluate the system under review and establish a system description or system model [43] [44]. This principle should apply to business systems as well as technical products [45]. An SMS is business systems engineering and therefore a description of its scope, functions, and context are critical to produce an appropriate SMS design. The proposed approach incorporates a few design principles at the system level. The following sections address where system safety principles may inform the design of the AV industry SMS.

---

[3] UL 4600 defines a safety case as "A structured argument, supported by a body of evidence, that provides a compelling, comprehensible, and valid case that a product is safe for a given application in a given environment."

**Value chain scope**

One lesson the AV industry can learn from aviation's experience outlined above is that an AV SMS framework should apply to the entire value chain from the outset. This is especially true since the range of business functions in the emerging AV sectors is broader than that found in the traditional automotive sector and is closer to the range of functions found in aviation. The range of aviation value chain functions that go beyond the scope of the typical automotive industry value chain include airport and air traffic operations, fleet maintenance and management, operational security, flight monitoring, and many other operating functions. As mentioned above, the AV industry has analogous operational considerations like ride-hailing, trucking, fleet management, as well as remote and in-field assistance, which should be included within the scope of the SMS.

**Range of elements**

Another fundamental question relates to the range of elements included in the SMS framework. Should the SMS only focus on the policies, promotion, and practices for managing and assuring risk, as we see in the aviation SMS model? Or should it provide guidance on generalized recommended practices for management processes which may pose risk if not performed well?

The majority of SMS frameworks include elements that go beyond the aviation model and provide guidance for management processes which may pose risk if not performed well. Given the novel and dynamic complexity of the AV industry, the authors believe this broader set of elements will provide better guidance on how to avoid foreseeable pitfalls as the industry evolves.

**Industry standards expand upon regulations**

As stated above, the regulatory environment for the AV industry is fragmented and still maturing. In this context, a reasonable concern can be raised about the direct use of the aviation SMS, which is built upon a robust pre-existing regulatory framework that does not exist in the AV sector, particularly for its operational aspects. Said another way, some regulated functions in aviation may not be regulated in the AV industry, but should be included in the SMS to provide safety management guidance. In aviation there are thousands of regulations which govern practices such as documents, records, training, parts control and traceability, operational control, and numerous other safety-related functions. No such regulatory guidance exists for the AV industry and should be considered as elements in the newly emerging AV SMS. Therefore, we should consider elements that are regulated in some industries to inform a robust AV SMS rather than simply carrying an SMS design from aviation to the AV industry. In fact, when we look at regulated functions in aviation that are not included in the SMS per se, we can often find them in other SMS frameworks.

As an example, The International Air Transport Association (IATA) established a series of Operational Safety Audit standards and recommended practices (ISARPs) which are used to augment the basic aviation SMS framework to support a more rigorous assessment of safety management. When comparing the four basic elements of the aviation SMS to the key management system processes found in the IOSA ISARPs, we find at least four major safety-related management functions that were added to IOSA to complement the core SMS framework. These include contracting and external service provider management, physical and cyber security, incident management, and management of change process [46].

The IOSA audit standard is used by many airlines globally to improve safety audit efficiency and continually update safety "standards to reflect regulatory revisions and best practices." [47]. This last statement implies, and is confirmed by the differences highlighted above, that the aviation SMS framework is not comprehensive and needs supplemental guidance on safety management practices in order to better address the needs of large global airlines that operate in a naturally complicated industry with substantial partnerships and extended enterprises. Considering the complexity of the emerging business models in the AV industry, we can infer that an SMS framework that is more robust than the minimum aviation SMS model would better align to the broader approach taken by IOSA.

*Integrating SMS and adjacent management frameworks*

An SMS is not the only management system a company may choose to implement. Quality, Project, Financial, Insurance, Enterprise Risk, and Information Security management systems are examples of management frameworks that are adjacent to an SMS. All of these business considerations and related management systems must coexist in a balanced manner with their natural tensions and synergies [4]. In fact, many of these management systems share some common functional elements, such as training, document and records, risk management, etc. Adding these additional elements helps management teams identify opportunities to improve implementation of the SMS.

This is true for two main reasons. First, a company does not need to develop redundant functional capabilities to serve these diverse management systems. A single training or document management capability is likely to be sufficient to serve various needs with minimal adaptation of requirements. Furthermore, if a company is already committed to one of the management systems highlighted above, it is likely that it already has some form of these capabilities which can be expanded to serve a broader user base. Second, where a company has not yet fully developed these management systems, they may 'divide and conquer' development of these capabilities to reduce the burden on a single organizational element. A capability refers to the combination of organizational, process, and technology aspects of business processes [48].

This approach can help companies integrate these functions and address conflicts that result from competing objectives. This approach can improve the consistency and cost effectiveness for development and maintenance of each of these management systems [29].

*Recognizing lexical ambiguity*

The English language is prone to confusing definitions and usage that can affect the logical and effective design of an SMS. One reference characterizes the issue this way: "Social and organizational factors such as leadership, authority, centralization, decision-making, motivation, mindfulness, stress, culture and even "safety" itself are grounded in concepts expressed in natural language with all of its ambiguities and imprecision." [28]. Two sources of ambiguity include:

- Polysemy, where "words with two or more related meanings", is one type of lexical conflict that can directly affect how an SMS hierarchical framework is designed and used [49]. For example, the Australian Institute of Health and Safety cites at least five common uses of the term 'risk' [50].
- Synonymy, where several different words may be used to describe the same or closely related concepts, poses similar lexical challenges when designing an SMS framework [51]. In the domain of safety, this can be seen in the inconsistent use of terms like hazard, risk, exposure, or cause.

These sources of ambiguity can be seen in various elements across different SMS frameworks as well. Looking at some of the SMS frameworks referenced above, we can find elements exhibit polysemy and synonymy. Terms such as governance, assurance, and oversight that may have overlapping meanings in one context or different meanings in another. Safety culture is a notoriously ambiguous term. Finally, we can find several ways of referring to the same concept, that while appear intuitively obvious, may still raise questions of the specifics that might be included. Consider the following examples:

- Documentation [36]
- Control of Documents [17]
- Documented Information [40]
- Data and Information Management [52]

When harmonizing elements into a common SMS framework, special attention should be given to resolving such ambiguity in a way that best serves the emerging AV industry, which may best be recognized as fundamentally an extension of the automotive industry. Therefore, a determination of the priority can be assigned based on that regulatory or industry standard precedent. This traceability can be used as a tie breaker for justifying the selection of one term or definition over another when harmonizing.

*Taxonomy development and validation criteria*

Taxonomies are intended to help organize diverse information into a common frame of reference. A taxonomy creates a hierarchy of information categories that structures content into a cascade of elements from broader, more abstract to more narrowly specific ones. Successful taxonomies organize complex information to make it more easily understood by users to help identify concepts within a context such as biology or safety. It follows then, that the organizing principles we use to design a taxonomy play a significant role in how well the resulting structure minimizes lexical ambiguity.

In the context of safety, the hierarchical structure of SMS functions must support diagnosis of complex business systems which may be contributing factors to organizational accidents. More detailed technical safety taxonomies also exist to standardize potential impacts, events, hazards, controls, and contributing factors [53]. Both SMS frameworks and more detailed technical safety taxonomies can benefit from reducing lexical ambiguity.

Several development and validation criteria can improve the usefulness of a taxonomy by driving a close evaluation of terms and definitions to reduce lexical ambiguity [54]. They include:

- Mutually Exclusive and Collectively Exhaustive (M.E.C.E): Terms and their definitions must not overlap, and there are no logical gaps for that set.
- Essentiality: An attribute or its qualities are so important as to be indispensable to the very nature of that set.
- Level of abstraction: The level of specificity or generality of an idea or element relative to the other elements in a set.
- 'Literary' and 'Use' warrant: Literary warrant justifies a term based on formal industry standards. Use warrant justifies a term based on common usage in a local community or specific industry or company's colloquial language.
- Diagnosticity: How well the term identifies symptoms of an underlying condition and potential for remediation.
- Usability: A means to test ease of navigation, decision consistency, and practicality.
- Internal and external validity: Internal validity is the alignment to a company's internal references. External validity is the alignment of the elements to industry norms.

The establishment of a new AV industry SMS requires a critical eye to taxonomic integrity. This is especially true given the industry is new and is a blend of three different industrial sectors with distinct safety vocabularies, regulatory environments, and cultural norms. These criteria can help to improve the logic, adoption, and use of an SMS design structure and limit design artifacts that impair their utility. Applying these criteria will also help avoid pitfalls of using structures from other industries that may not be applicable to the AV industry.

*Process for comparison and harmonization*

The safety management frameworks referenced above have elements often placed at different levels across their hierarchies. As such, efforts to develop a more robust AV SMS need a way to compare and contrast these different frameworks to allow harmonization. This was accomplished by compiling the table of contents of each framework into a centralized catalog. The list of approximately 1,700 elements was then analyzed for redundancy, which implied importance, and variability which implied lexical ambiguity.

For example, Safety Assurance is found in both the ICAO SMS [7] and in API RP 1173 [17]. Additionally, close equivalencies in content and definition were found in the concept of record keeping where the same element appears at different tiers within different hierarchies. There are also some elements that in some frameworks are treated as core functions written once with the expectation that they are to be applied in all other relevant elements, while in other frameworks they are repeated within key elements and cited at the appropriate process step to reinforce their importance. Continual Improvement is an example of how an element is treated differently across frameworks [17] [40].

The final step in this process was to apply the taxonomy development and validation criteria cited above, and to sort and organize a harmonized hierarchy from the top down. This process required significant trial and error to negotiate terms and definitions in order to resolve inherent challenges resulting from the diverse lexicon of the references frameworks. During this exercise, it became clear that defining ontological characteristics for the taxonomic 'buckets' would aid in the harmonization effort. More work could be done in this area. The following is an initial proposal and that we expect it to be improved over time with feedback.

## Proposal

It is critical at this juncture to reiterate what this proposal is intended to achieve. This proposal is intended to:

- Provide expansive and robust voluntary SMS guidance.
- Improve the diagnosis of safety management functions that are likely to contribute to safety loss.
- Provide a comprehensive yet generalized and scalable set of management functions which can be easily understood by industry stakeholders.
- Enable integration into existing company management processes rather than act as a separate framework.
- Be applied across the entire value chain of a newly emerging AV industry.
- Provide context for and enable reference to other relevant and applicable industry safety and technical standards.
- Drive a continual improvement discussion by providing thought provoking potential solutions to long standing concerns about various SMS frameworks.

### *Proposed AV industry SMS framework*

As a result of the analysis of the referenced SMS framework, we propose the following enhanced SMS model for the AV industry, building on the currently accepted SMS models outlined above:

1. **Safety Culture Management**: This proposed element includes a closed-loop structure for the management of safety culture loosely based on Edwards Deming's Plan-Do-Study-Act (P-D-S-A) cycle [55]:

    - Set expectations: Establish and communicate expectations for what a positive safety culture is, and how leaders and individuals are expected to act. This includes relevant values, behaviors, and investments.
    - Enable adoption: Support leaders' and individuals' ability to demonstrate expectations across the organization with techniques such as training, regular communications and promotion, coaching and mentoring, and leadership reinforcement.
    - Monitor progress: Conduct activities to monitor and assess the effective follow-through and performance to established expectations, using such techniques as surveys, audits, leader-in-the-field assessments.
    - Make Improvements: Implement necessary corrective and preventive actions to mitigate factors impeding fulfillment of expectations and to preserve a strong safety culture. Examples may include updated policies and procedures, new or revised expectations, enablement, or monitoring methods, as well as implementation of performance management techniques for leaders and individuals who are unable to meet the established expectations.

2. **Safety Planning**: Planning bridges leadership's intent to maintain a desired safety culture and its governance of the organization's efforts to achieve an effective SMS. The outputs of planning manifest a company's safety objectives, strategies, policies, and priorities into resourced and actionable programs and management practices. These programs and practices should effectively enable the organization to carry out the comprehensive set of day-to-day safety activities associated with the SMS. Effective safety planning should include the following elements:

    - Safety Policy and Objectives: Establishment and maintenance of processes to ensure the company can articulate and manage plans which impact safety. These include strategic planning, setting safety objectives, establishing safety policies, organizational design considerations, and annual planning.
    - Compliance Management: Establishment and maintenance of processes to ensure the company is able to manage the life cycle of compliance requirements, including identification of requirements relevant to the company, determine applicability of those requirements by function, product, and location, assess and close gaps, assess effectiveness of compliance methods developed through assurance activities, compliance reporting, as well as monitor for new and changing requirements.
    - Information Management: Establishment and maintenance of processes for effective management of documents, records, and information security.
    - Training and Competency Management: Establishment and maintenance of processes necessary to effectively manage the development and delivery of safety sensitive knowledge, skills, and abilities for personnel.
    - Project and Program Management: Establishment and maintenance of processes for effective management the scope, resource allocation, scheduling, stakeholders, dependencies, and status reporting to assure the safety implications of initiatives are addressed in an efficient manner.

3. **Safety Governance:** Governance is the higher order function that directs effective implementation of safety plans and the overall SMS. Safety Governance is intended to be the element that provides leaders, regulators, and the public with confidence that the company is successfully managing the business of safety. Safety Governance includes key functions such as:

- <u>Safety committees</u>: Establishment of standing and ad hoc review and decision-making bodies tasked with determining safety and deployment readiness. This responsibility extends from maintenance of safety methodologies to the most significant safety decisions at the executive level. This layered, multi-disciplinary approach provides important checks and balances involving safety subject matter experts who are independent of engineering, product development and operations, and is a key factor to successful safety management.
- <u>Safety concern reporting</u>: Effective governance depends on leaders being aware of the true status of how well safety is being managed. While data driven analysis is essential, the frontline workforce is able to detect off-nominal conditions which cannot be found in the data [56]. Systems such as NASA's Aviation Safety Reporting System show how employee reporting is a complementary source of system monitoring to improve the richness of data needed to effectively govern safety [57].
- <u>Safety assurance</u>: The layers of complementary internal, independent, and external assessment and evaluation activities, such as compliance assurance, inspections, as well as self-assessments, internal and external audits. One important artifact of safety assurance activities is the Safety Case. A safety case for fully autonomous operations is a formal way to explain how a company determines that an AV system is safe enough to be deployed on public roads without a human driver, and it includes evidence to support that determination [58].
- <u>Contracted services safety management</u>: Processes to manage the relationship between strategic business partners, vendors, contractors and other 3rd-party contracted service entities to ensure safety objectives are met.
- <u>Safety performance monitoring</u>: Focuses on the safety performance of products and services a company develops and delivers, as well as the business processes associated with the SMS itself. This element includes requirements for data collection, reporting, analysis, and assessment of information related to the how well safety is being managed, as well as the regular leadership assessment of the design adequacy, conformance, resilience, and performance effectiveness of internal controls established by the SMS risk control expectations established in safety objectives.
- <u>Interface management</u>: A repeatable set of processes intended to address an acknowledged but heretofore unstructured approach to improving governance of interfacing organizational activities, objectives, incentives, dependencies, and constraints.

4. **Safety Risk Management**: The broad set of functions used to manage different aspects of safety risk proactively and responsively at the enterprise level.

- <u>System Safety</u>: The proactive management of safety risk, system safety is a discipline that applies engineering and management principles, criteria, and techniques to achieve an acceptable or reasonable level of risk throughout the product or services life cycle.
- <u>Emergency Preparedness</u>: Processes include planning, risk identification and assessment, impact analysis, response and recovery planning, exercise and testing, and improvement.
- <u>Event Investigation</u>: Processes associated with effective evaluation of causal factors related to safety events after the fact, including preparation, coordination, analysis, documentation, monitoring actions, and lessons learned.
- <u>Change Risk Management</u>: Management of change drivers that could potentially impact safety, such as changes to regulations, products or services, or goals and objectives. Relevant processes are used to identify sources and triggers, determine applicability and impact, development and implementation of plans, and the validation and monitoring of plan implementation.
- <u>Workforce Involvement</u>: Processes used to assess the completeness and validity of all safety risk management activities, whether proactive or responsive (reactive). Sample processes include identification of relevant personnel, seeking and assessing input, supporting implementation, and sharing successes.

5. **Safety Life Cycle Management:** The range of safety-related functions which are the core of the value chain. This element highlights recommended core product life-cycle practices which may have an impact on safety and includes the following sub-elements.

- <u>Engineering</u>: A broad set of topics such as design and development of the system after rigorous risk and hazard analysis, test and validation of the system and subsystem components, monitoring AVs continuously after deployment to identify new risks and hazards, establishing a procedure for incident reporting and analysis,
- <u>Manufacturing</u>: Processes to train personnel on manufacturing processes and safety protocols, implement rigorous quality control to production processes, ensure externally sourced components meet safety and quality standards, document all processes to ensure traceability, and modify manufacturing processes based on feedback, changes in requirements, or changes in safety standards.
- <u>Operations</u>: Operational safety control protocols, routine maintenance and updates to hardware and software components of the system, standardized training and certification of personnel, remote and in-field assistance, first responder interactions, user engagement and education, and ensuring all procedures are compliant with regulatory standards are covered in this element.
- <u>Disposition</u>: Planning for decommissioning including timeline, procedures, responsible parties, methods to properly identify and handle hazardous materials, establishing processes to reuse and recycle vehicle components, if possible, documentation of end-of-life and disposal process, assessment of impact on environment aiming for minimal ecological footprint, and communicate relevant information and details to stakeholders.

## Notable inclusions

This section outlines the areas where significant additions have been made to basic transportation SMS models and provides justification for their being included in the proposed AV SMS framework.

**Safety Culture Management**: In the context of any SMS, safety culture is seen as a fundamental principle. Safety culture is the backdrop for all other SMS activities and informs companies' SMS design and implementation. Although safety culture is referenced in many safety management frameworks in a cursory manner, the work of INPO, CCPS, and NASA, are notable for standardizing this element in depth [31] [32] [59]. Safety culture is most rigorous in these industries because of the high degree of public scrutiny for catastrophic accidents. While ADS-equipped vehicle accidents may not cause large scale safety loss in a single incident by being local collisions, the centralized nature of design and operational control may create a broad sense of public exposure to potential safety incidents. Therefore, there is a need for at least a basic, closed loop management approach to manage safety culture. The framework proposed here is relatively simple and leverages existing and basic cultural and Human Resources management practices, such as establishing leadership competencies that can help improve a company's approach to Safety Culture, while not being overly prescriptive about what their culture should look like [60]. Although Safety Culture is portrayed as a crucial component in the SMS framework, this document does not yet provide detailed information on the characteristics or traits of a healthy safety culture. Future efforts may include the development of such principles.

**Project and Program Management:** Effective management of development of and updates to a company's products and services can have an impact on safety [61]. Project and program management is integral with the functions of Change Risk Management [62] which identify, assess, and implement new and changing requirements. However, project and program management add unique value in an SMS by helping to manage resources, schedules, and scope. Additionally, and perhaps most critical to effective safety management, project management techniques help identify critical dependencies and project sequencing, which can help to improve visibility of how safety can be brought in earlier in a project's life cycle to increase efficiency. A company that is not effective in understanding the dependency of safety in the sequencing of the critical path will likely find negative impacts to cost, time, or quality. And while companies will make every effort to place safety ahead of these other factors, history shows that the cultural pressures on production can be overwhelming and may ultimately negatively impact safety. Therefore, this framework includes project management as an important function to effectively help manage safety.

**Contracted Services Management:** Managing an extended supply chain and outsourcing of business functions are common practices. However, while a company "can outsource the work, ... you can't outsource the risk" [63], and coordination of vendor relationships is both complex and dynamic. The implications for the quality of a company's products and services have been widely held to be a critical function in many referenced non-aviation safety and quality management frameworks. It is also important to note that even aviation IOSA ISARPs include management of contracted services [46]. Additionally, the UNECE established requirements for ADS manufacturers to establish processes to evaluate supplier SMS capability [42]. Therefore, this function has already been included in the proposed SMS framework within the AV industry.

**Interface Management:** This function is based on interpretation of the extensive research on organizational and management factors that contribute to safety vulnerabilities [4]. The SAE ORAD SMS Task Force has attempted to translate this research and direct observation of the safety vulnerabilities caused by internal and external organizational silos. By applying a generic closed-loop management process, the Task Force formulated a generalized set of management activities that can serve as a model to identify and mitigate these pervasive management challenges.

**Safety Life Cycle Management:** These safety-related management functions are often regulated in many industries such as aviation. Yet some safety management frameworks include these functions as a core element [52]. However, as discussed above, when designing a new system, including an SMS, the context of the new system must be understood and taken into account. Just because a safety-related management function is regulated in one industry does not mean it should be excluded from another industry's SMS. This is particularly true when the other industry is newly forming, with an emerging regulatory framework, such as what is found in the AV industry. Therefore, this element is an attempt to extract key safety management processes that may be regulated in some industries or included in industry SMS standards because they can influence safety outcomes. This set of elements also includes key safety management practices unique to the AV industry, such as first responder interactions and remote and in-field assistance.

## Notable changes in hierarchy

This section outlines significant changes in hierarchical structure that have been made to improve the utility and logic of the taxonomy, and provides justification for the changes.

**Safety Policy and Objectives (SPO):** This element was brought to a lower level in the hierarchy under the broader topic of Safety Planning. Rationale for this change is seen in the relationship of new peer functions, like Information and Training and Proficiency Management. This change is a clear example of the application of the 'Level of Abstraction' criteria to the taxonomy. If Safety Policy and Objectives were to be retained at the top level, the other elements contained within the proposed Safety Planning set would be an awkward fit. In the current proposal, there are five functions that essentially serve as preparation or planning to govern, with establishment of safety policies and objectives being the first among them.

**Safety Governance vs Safety Assurance (SA):** Different terms for this function are used in different industry standards. In the most granular example, audit or management review can be found, yet are more tactical activities than intended for this broader and more inclusive management function. At the next higher level of abstraction, Safety Assurance found in many existing SMS frameworks includes safety performance monitoring activities, management of change, and continual improvement. However, conceptually 'Governance' can generally be viewed as a higher order function than 'Assurance'. Logically, assurance can better be viewed as one mechanism within a broader governance function. Finally, the term 'safety oversight' has been proposed as an alternative but is generally more closely associated with governmental or regulatory inspection and compliance activities, rather than a company's internal functions. As a result of this analysis and in the context of the AV SMS, Safety Governance was deemed the most appropriate term to provide the right level of abstraction to include activities intended to maintain a high level of confidence that an SMS is successful at delivering the absence of unreasonable risk in the AV value chain.

**Safety Risk Management (SRM):** A review of many existing frameworks reveals that SRM carries a direct functional equivalency to System Safety methodologies [64] [44]. This presents two main problems. The first is the direct overlap of the general term "safety risk management" and the formal discipline of System Safety or System Safety Engineering, which promotes ambiguity. Over time, SRM has been conflated to mean System Safety Engineering, as applied to the proactive design of products and services. The second problem results from this overlapping definition and is that this narrower definition of SRM tends to exclude other enterprise-level safety risk management functions, such as Emergency Management, which must then find a home in another element. This proposal increases the abstraction of "safety risk" to include a broader set of enterprise functions.

**Emergency Preparedness:** This element is found as a broad requirement under Safety Policy or in the Planning element of various SMS frameworks. This function is essentially the reactive management of safety risk which complements the proactive approach found in System Safety and should therefore be included under the newly broadened Safety Risk Management element. Additionally, the proposed AV SMS framework outlines a basic management process which strongly echoes that of System Safety, further strengthening alignment of these two elements.

**Workforce Involvement**: In the context of the SMS, the essential purpose of workforce involvement in an SMS is to ensure the organization obtains:

- A comprehensive understanding of the types and nature of hazards that actually or potentially exist.
- The full range of potential controls or mitigations are available.
- The viability or efficacy of those controls or mitigations.

In terms of safety, workforce involvement is a critical method to validate both system safety and emergency management practices. This element is intended to operationalize within the SMS, the concepts often referred to as "work-as-imagined and work-as-done."

[65]. It is for this reason that the current proposal places this function within the risk management element; it is a reminder of the importance of fully engaging front line staff to understand how and when to enhance management of safety risk, and not simply invite employees to town halls and be the recipients of safety communications.

## *Challenges*

This section describes potential points of resistance to the changes proposed here, as well as rationale for defense against these objections.

**Adoption of the aviation SMS model:** This proposal departs from the frameworks proposed by AVSC, UNECE, and the EU, which are consistent with the aviation-based SMS framework promulgated by ICAO, FAA, and the broader transportation community. However, a basic review of Wikipedia reveals several other industry sector standards that provide additional safety management insights [66]. A deeper review reveals many more expansive SMS frameworks. Extensive research into these frameworks has revealed a broader set of important elements missing in the aviation-based SMS. Therefore, given the context of the newly emerging AV industry and the opportunity to design a robust SMS, this approach stands as a complement to the product established by the AVSC and others.

In fact, as demonstrated by the complementary arrangement between the aviation SMS and the IOSA ISARPs, it is possible for both the current and the proposed AV SMS frameworks to co-exist. The current AV SMS frameworks can stand as a minimum viable product for the purposes of regulatory requirements and formal certification, and the framework proposed here can serve as an advanced framework to help companies strive for industry-leading safety practices.

**Concerns about over-reach:** It is common practice for many industry safety management frameworks to include a broad set of functions. In fact, this breadth of functions has been included here specifically because they may have been identified as contributing factors to accidents. It is the duty of SMS designers to learn from the experience of other industries and be proactive in their adoption. Ultimately, safety best practices recognize the efficiency of early, voluntary adoption to avoid more costly and less flexible mandatory implementation of safety requirements. Additionally, assuming this framework is not mandated, it offers companies a rich menu to address problems they identified themselves.

**Implementation, and assurance guidance:** The proposal currently only provides the framework of what functions should be included in a robust AV industry SMS. Reviewers of the proposal are likely to ask which elements are best to implement first, or in what particular sequence should they be implemented. Additionally, questions are likely to arise about how a company knows they have achieved a sufficient level of effectiveness, i.e., what does good look like?

Given the early stages of the AV industry and the broad set of functions found in the proposed framework, questions about implementation and assurance will be addressed in future efforts. This approach follows the example established by the FAA, which published three separate documents outlining the framework, implementation guidance, and assurance guidance for the pilot project participants in the early voluntary implementation phase of its SMS programs [67] [68].

## Conclusion

This proposed SMS framework was developed for the emerging AV industry. The new AV industry encompasses an expansive life cycle and integrates three disparate legacy industry sectors within a dynamic and maturing regulatory environment. Initial efforts to define an AV industry SMS relied heavily on aviation's model. However, other industries have established more comprehensive safety management frameworks which predate aviation's efforts and should also be considered in the design.

A comparative analysis of various additional safety management frameworks was conducted to catalog potentially important elements. This compilation was then subjected to a harmonization effort, which applied taxonomy development and validation criteria to create a more complete and durable AV industry safety management framework.

While the current aviation-based AV SMS framework can stand as a minimum viable product for the purposes of regulatory requirements and formal certification, the framework proposed here is intended to serve as an advanced framework to help companies strive for robust, industry-leading safety management practices.

## Acknowledgments

The authors would like to thank Dan Bartz, Chris Bartholomew, Matt Schwall, Mauricio Peña, Trent Victor, and Ed Straub for their careful reading, comments, and insightful suggestions.

## Appendix A

- Asset Management: ISO [69], IAM [52]
- Automotive: AVSC [34], ISO [36], [37], ISO/IATF [40], EU [2], UNECE, [42], ANSI/UL [38]
- Aviation: ICAO [7], FAA [11], IOSA [46],
- Chemical manufacturing: CCPS [30], [32]
- Department of Defense: MIL-STD 882E [64]
- Integrated safety: Bahr [44], ISO [29]
- Maritime: IMO ISM [70]
- Nuclear: INPO [31], IAEA [71], [72]
- Occupational: ISO [73], ANSI/ASSP- [74]
- Pipeline: API - RP1173 [17]
- Rail: FRA [26], Japan Rail [18]
- Transit: FTA [25]

Developing a Safety Management System for the Autonomous Vehicle Industry